\documentclass[eat,twocolumn]{jmlr}
\usepackage{amsmath}
\usepackage{amssymb}

\newcommand{\R}{\mathbb{R}}

\usepackage{longtable}% for long tables

\usepackage{booktabs}
\usepackage[load-configurations=version-1]{siunitx} % newer version
\theorembodyfont{\upshape}
\theoremheaderfont{\scshape}
\theorempostheader{:}
\theoremsep{\newline}

\jmlrvolume{}
\firstpageno{1}
\jmlryear{2022}
\jmlrworkshop{Machine Learning for Health (ML4H) 2022}

\title[Automatic Infectious Disease Classification Analysis with Concept Discovery]{Interpretability of Automatic Infectious Disease Classification Analysis with Concept Discovery}

  \author{\footnotesize Elena Sizikova, Joshua Vendrow, Xu Cao, Rachel Grotheer, Jamie Haddock, Lara Kassab, Alona Kryshchenko, Thomas Merkh, R. W. M. A. Madushani, Kenny Moise, Annie Ulichney, Huy V. Vo, Chuntian Wang, Megan Coffee, Kathryn Leonard, Deanna Needell}

\begin{document}

\maketitle

\begin{abstract}
Automatic infectious disease classification from images can facilitate needed medical diagnoses. Such an approach can identify diseases, like tuberculosis, which remain under-diagnosed due to resource constraints and also novel and emerging diseases, like monkeypox, which clinicians have little experience or acumen in diagnosing. Avoiding missed or delayed diagnoses would prevent further transmission and improve clinical outcomes. In order to understand and trust neural network predictions, analysis of learned representations is necessary. In this work, we argue that automatic discovery of concepts, i.e., human interpretable attributes, allows for a deep understanding of learned information in medical image analysis tasks, generalizing beyond the training labels or protocols. We provide an overview of existing concept discovery approaches in medical image and computer vision communities, and evaluate representative methods on tuberculosis (TB) prediction and monkeypox prediction tasks. Finally, we propose NMFx, a general NMF formulation of interpretability by concept discovery that works in a unified way in unsupervised, weakly supervised, and supervised scenarios\footnote{Our code is available at: \url{https://github.com/esizikova/infectious_disease_concepts}}.
\end{abstract}

\section{Introduction}
%goal 
Visualization of learned features is crucial for a better understanding of patterns learned by neural networks. Concept discovery approaches are a type of explainable artificial intelligence (XAI) technique that identifies high-level and human-understandable explanations for the predictive behavior of a model. In this work, we analyze infectious disease classification neural networks using concept discovery approaches.  
 
 % challenge
While neural networks achieve impressive results, they mostly remain black-box models. Visualization and interpretation of learned representations remains an important challenge in their analysis~\citep{lou2012intelligible}, and is one of the main hurdles to wider adoption of data mining and AI technology for many applications~\citep{kim2017interpretable,bussmann2020explainable,islam2021explainable}. Specifically for healthcare~\citep{singh2020explainable}, explainability is crucial for analysis of success and failure cases that is typically done by medical professionals. There have been instances where AI tools have been found to determine classifications based on meta or extraneous data, which would not be reliable when repeated in medical contexts. Other tools are simply black boxes. The contexts in which and the patients for which these tools may fail matter. Biological plausibility is a key factor in determining whether an intervention will be adopted in medicine. It is important that a diagnosis is based not on a correlation that may change but on a substantive basis. AI tools may be biased and may not perform as well in certain populations, perhaps related to race, gender, age, or other characteristics, depending on the composition of training datasets. To the clinician, what matters is the patient in front of them, not the average performance. Medical ethics requires physicians to ``do no harm". Medical legal liability requires clinicians to be responsible for the decisions they make with the use of AI tools. Many AI tools are developed and deployed without Randomized Controlled Trials to demonstrate replicable results across a broad array of populations and sub-populations. Other interventions in medicine require substantial validation in a real world setting. What works in the lab or in a specific population may not work in the real world clinical situations where clinicians work. The impact can be devastating in medicine if clinical diagnosis is affected; clinicians may even learn to rely on tools and be deskilled, leading to worsened outcomes than would result without the tool. In medicine, not all that seems good is in fact good. What may seem like a step forward can inadvertently be a step back.

% approach
Many XAI techniques require a particular network architecture, access to network weights, and back-propagation to generate an interpretation heat map~\citep{chattopadhay2018grad,zhou2016learning}. Additionally, it is crucial to link generated explanations to existing human knowledge. Therefore, for studying infectious disease classification models, we seek XAI methods that are model-agnostic, fast, provide a global overview of model behavior, and generate human interpretable outputs. Concept extraction (CE) approaches are a class of XAI techniques that seek to explain the decision-making process of a neural network in human-understandable terms, or so-called concepts. While gaining popularity in the image analysis community, CE approaches are not widely used for analyzing medical image classification models. In this work, we seek to explain the behavior of neural network (NN) infectious disease classification models using post-hoc visual explanations generated by CE methodology proposed in \citep{collins2018deep, oramas2019visual, posadamoreno2022eclad} (we provide an extensive discussion of related work in the Appendix). As a result we can visualize information encoded in features and analyze whether encoded topics consistently capture image regions that are semantically related. In comparison to existing techniques, the presented method is flexible to work in unsupervised, semi-supervised or weakly supervised fashion, and provided labels do not need to correspond to the labels that the underlying network was trained for.  The extracted concepts provide a useful visualization tool to medical image professionals, and match the intuition of where doctors would search an image for the presence of disease. In summary, our contributions are as follows: (1) To understand infectious disease classification behavior, we propose to extract and analyze concepts generated by CE approaches on image datasets for tuberculosis (TB) and monkeypox, both infectious diseases that are prone to under-diagnosis and require prompt identification for treatment and to prevent others being infected. (2) We compare concepts generated by the feature factorization methods proposed by \citep{oramas2019visual}, \citep{posadamoreno2022eclad} and \citep{collins2018deep}, extending the latter to accept weak supervision (NMFx framework). We study medical image datasets (TBX11K~\citep{liu2020rethinking}, monkeypox~\citep{moise22monkeypox}) as well as PASCAL-VOC 2010~\citep{pascal-voc-2010}, a public image analysis dataset.

\section{Methodology}

\begin{figure*}[t]
    \centering
    \includegraphics[width=0.9\linewidth]{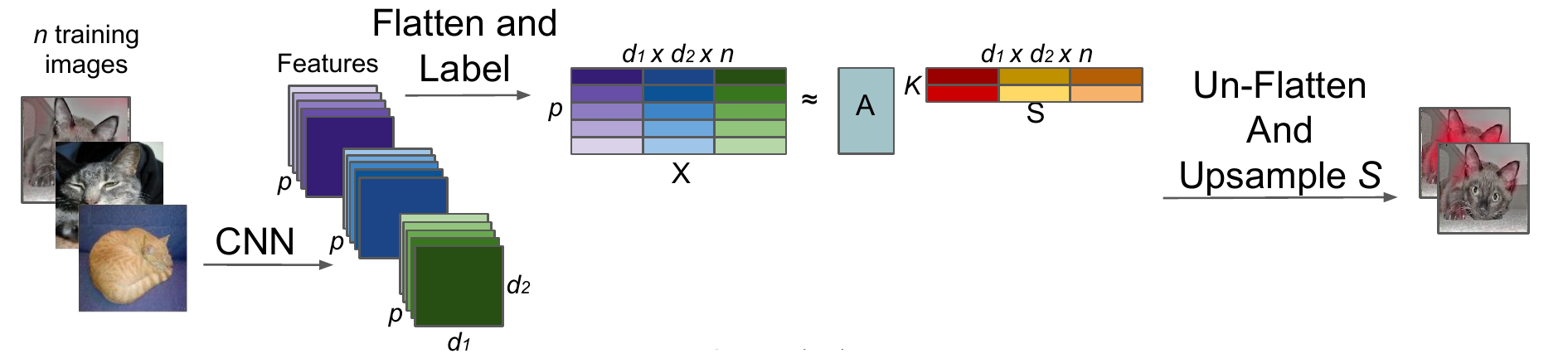}
    \caption{\emph{Visual explanation of NMFx. Optional image labels can be used during the modified optimization step, (SS)NMFx. Example above is shown for $K=2$ and $n=3$.}}
    \label{fig:vis_reshaping_overview}
\end{figure*}

% \subsection{Concepts Extracted Using NMF.} \label{sec:factorizationintotopics}
The first method we consider is proposed in ~\citep{collins2018deep}. The general idea is to extract feature representations of images using a convolutional neural network and factorize the resulting reshaped matrix using NMF with $K$ topics. After a reshaping procedure, the weight matrix results in a set of $K$ heat maps that visually explain information encoded in the features. The process is shown in Figure~\ref{fig:vis_reshaping_overview}.

Let $X \in \R^{n_{1} \times n_{2}}$ denote the nonnegative data matrix of $n_2$ data points in $\R^{n_1}$. Lee and Seung~\citep{lee1999learning}  propose to decompose X into a topic matrix $A$ and a weight matrix $S$ using the following, Frobenius-norm optimization objective:
\begin{equation}\label{eq:NMF}
\min_{A,S}    \|X - AS\|_F^2.
\end{equation}
Here, $A \in \R^{n_1 \times k}_{\ge 0}$ denotes the topic matrix with $k$ topics and $S \in \R^{k \times n_2}_{\ge 0}$ denotes the representative weight matrix. 

\noindent \textbf{NMF with Image Label Supervision}
When information about the data points' labels is available, we can encode it into $Y \in \R^{l \times n_{2}}$, a binary label matrix where columns correspond to data points in $X$ and rows represent their class membership. Lee et al.~\citep{lee2009semi} then propose the classical semi-supervised nonnegative factorization (SSNMF) method, whose objective is given by:
\begin{equation}\label{eq:SSNMF}
 \min_{A,B,S}   \|X - AS\|_F^2 + \lambda \|Y - BS\|_F^2, 
\end{equation}
where $\lambda \in \R$ is a regularization parameter and $B \in \R^{l \times k}$ is the trained classification matrix. The first term in objective (\ref{eq:SSNMF}) denotes the reconstruction error of the factorization and the second term denotes the classification error. This model simultaneously learns a topic model, defined by the matrix $A$, and a classification model, defined by the matrix $B$.  The matrix $S$ provides a representative weight matrix which both fits the topic model and predicts labels. We refer to the resulting generalized concept extraction technique as NMFx.

\section{Implementation Details}
\subsection{NMF.} 
Given $n$ images, we obtain their network feature representation $X'$ with dimension $(n,p,d_1,d_2)$, where $p$ is the number of feature maps, and $d_1,d_2$ are the width and height of each feature map, respectively. For example, using the features after the rectified linear unit (ReLU) and the last convolutional layer of VGG-16~\citep{simonyan2014very}, we have $p=512$ and $d_1=14$ and $d_2=14$. $X'$ is then flattened and transposed into matrix $X$ of dimension $(p, n\times d_1\times d_2)$, before being passed to the NMF or SSNMF optimization objective, and obtaining the factorization $X\approx AS$ with $K$ topics. Note that in this setting, a data point is a $p$-dimensional vector representing a location in an image. The resulting nonnegative weight matrix $S$ is of dimensionality $(K,n \times d_1 \times d_2)$ and, if a train-test split is used, we obtain $S_{\text{test}}$ of dimension $(K,n_{\text{test}} \times d_1 \times d_2)$ using nonnegative least squares, as described above. We reshape $S$ into a heat map tensor of dimension $(n,K,d_1,d_2)$ and up-sample to image resolution $(n,K,w,h)$,  where $w$ and $h$ are the width and height of the input images, respectively. The reshaping procedure for $S_{\text{test}}$ is analogous to that of $S$. 

Whenever the SSNMF is used as the objective for NMF, we also create a binary class tensor $Y'$ of dimensionality $(n,K,d_1,d_2)$. For each image $i$ of label $l$, the $d_1\times d_2$ sub-tensor corresponding to this image and label are set to $1$, otherwise it is set to $0$. Subsequently, $Y'$ is reshaped and transposed to matrix Y of dimensionality $(K, n\times d_1 \times d_2)$ and used in the optimization.

\section{Datasets}
We evaluate concept discovery using NMFx and other techniques on TBX11K~\citep{liu2020rethinking}, a public tuberculosis chest X-ray dataset. We rely on the official data split in \citep{liu2020rethinking} that includes multiple smaller, public sets~\citep{jaeger2014two}. We also evaluate our approach on the task of monkeypox classification. For this task, we consider a subset of 356 monkeypox and 345 non-monkepox images from publicly available images in the medical literature and from social media and journalistic sources~\citep{moise22monkeypox}. These images include confirmed diagnoses. Monkeypox images were all from clade II and only included if documented as confirmed by PCR (polymerase chain reaction) testing. Monkeypox images were included from all stages of lesion evolution. Comparison images were selected by medical doctors as appearing similar to monkeypox and having similar clinical syndromes (such as herpes, syphilis, varicella, hand foot and mouth disease, and molluscum). Some images were also identified by google image searches for monkeypox images to identify similar appearing images. These images were cropped to include only the skin lesions and eliminate identifying information (such as jewelry, tattoos, facial features, and clothing) which could affect results and to ensure images were closely matched. The monkeypox and non-monkeypox image datasets were compared to show similar proportions by age (adult or child), skin type, sex and gender, and body part affected. 

\section{Results}
We now analyze the proposed technique and its variations.

\begin{figure*}[ht]
    \centering
    \includegraphics[width=\linewidth]{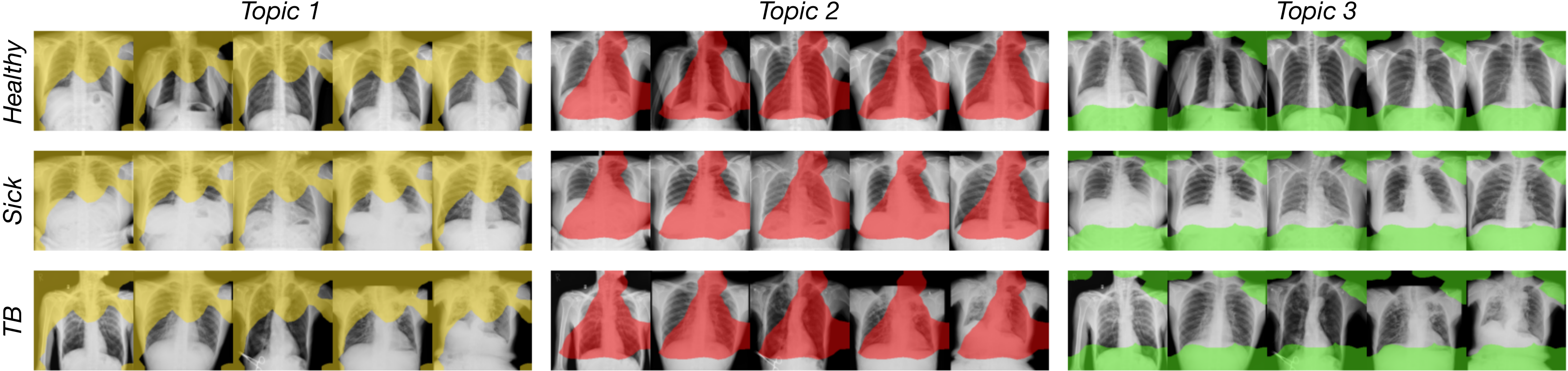}
    \caption{\emph{Visual explanations using NMFx for a VGG-16 Tuberculosis classification task.} Clinically, Topic 1 corresponds to areas used most in diagnosis, Topic 2 corresponds to areas sometimes involved in diagnosis, and Topic 3 corresponds to unrelated areas. Colors are chosen randomly.}
    \label{fig:vis_tbx_nmfx}
\end{figure*}

\paragraph{Visual Topics Found in Tuberculosis Analysis}
In Figure~\ref{fig:vis_tbx_nmfx}, we visualize topics found using NMFx in the VGG-16 network trained for a tuberculosis classification task. The colors were arbitrarily chosen. In particular, topic 1 (yellow color) corresponds to the areas of interest corresponding to predicting tuberculosis. Tuberculosis disproportionately affects the upper lung fields, which are highlighted in yellow, unlike other infections, which makes this area very important in identifying tuberculosis. Topic 2 (red) highlights  areas that are less important, but can be involved, in tuberculosis diagnosis, such as the lower lung fields (potentially has dependent pleural effusions) or central sections (potentially has pericardial effusion or hilar lymphadenopathy). The Topic 3 (green) corresponds to areas outside the lung fields which are not expected to be helpful in TB diagnosis. We also find that visual topics consistently highlight similar anatomic areas across a variety of input example anatomies.

\begin{figure*}[ht]
    \centering
    \includegraphics[width=0.9\linewidth]{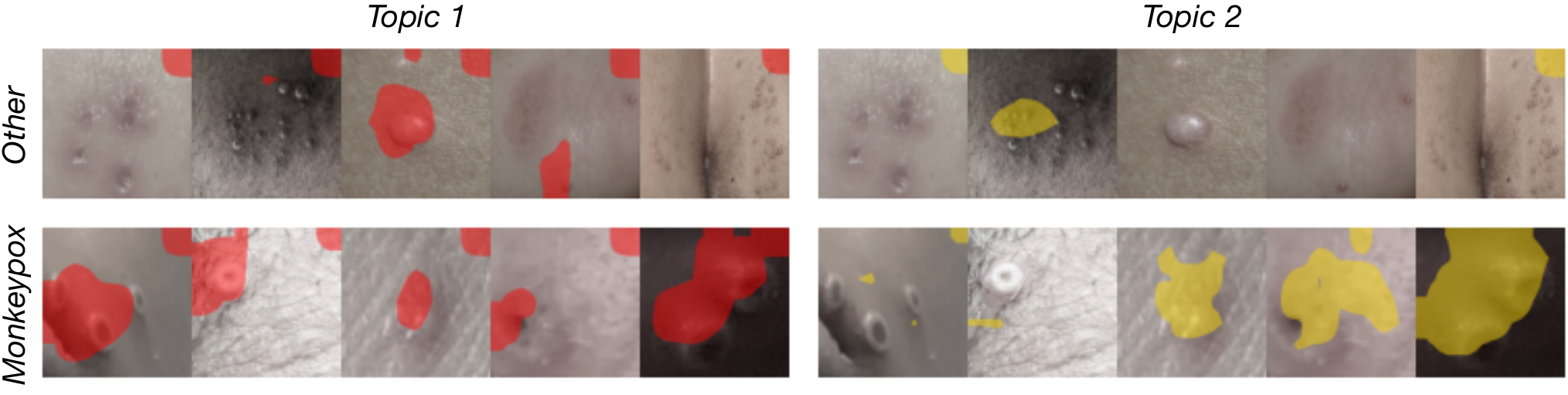}
    \caption{\emph{Visual explanations using NMFx for a monkeypox classification task.}}
    \label{fig:vis_monkeypox_nmfx}
\end{figure*}

\paragraph{Visual Topics Found in Monkeypox Analysis}
In Figure~\ref{fig:vis_monkeypox_nmfx}, we look at the topics found by the EfficientNet-B3 network trained for the monkeypox classification task. We find that the topics are centered on the lesions and regions corresponding to monkeypox lesions (second row). On the other hand, in examples with non-monkeypox skin conditions (first row), the topics identify some visually similar lesions but are more scattered and occupy a smaller surface area. 

\begin{figure*}[]
    \centering
    \includegraphics[width=\linewidth]{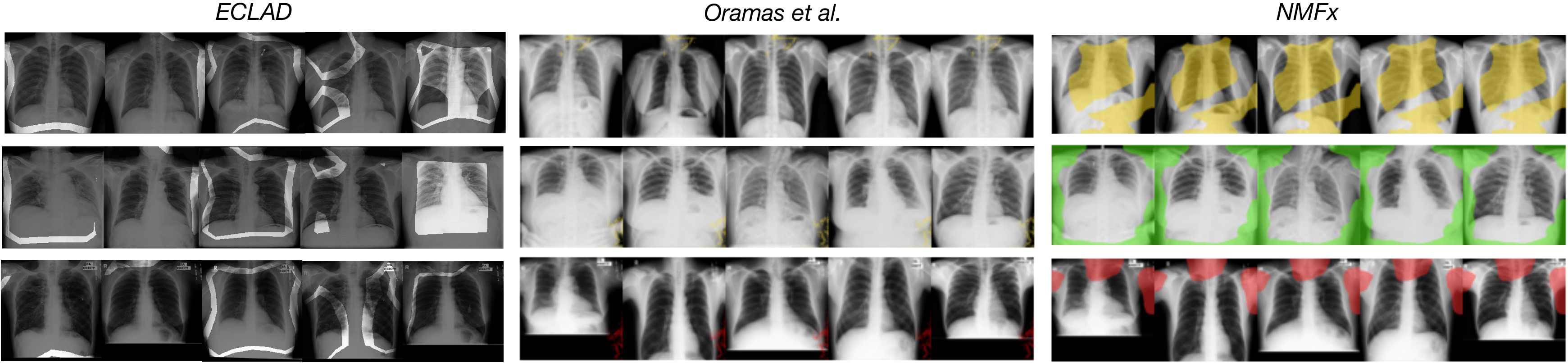}
    \caption{\emph{Comparison of visual concepts identified by different techniques.}}
    \label{fig:vis_comparison}
\end{figure*}

\subsection{Comparison to Other Automatic Concept Extraction Techniques} \label{sec:factorizationoramas}
We compare the above approach to two other, representative automatic concept extraction methods: Oramas~\citep{oramas2019visual} and ECLAD~\citep{posadamoreno2022eclad} in Figure~\ref{fig:vis_comparison}. Please see Appendix for additional implementation details details. For this experiment, we consider approximately 30\% of the the TBX11K~\cite{liu2020rethinking} dataset, due to  computational limitations of running \cite{collins2018deep,posadamoreno2022eclad}. In contrast to ECLAD and the method of Oramas, the NMFx method identifies larger and more consistently positioned regions in the input X-ray images.

\section{Conclusion}

In this work, we explore three automatic concept discovery approaches as explainability tools for neural networks trained to classify infectious diseases. We find that generated concepts highlight areas where disease lesions are generally found (as in tuberculosis) or noted in specific, available images (monkeypox). The highlighted areas allow clinicians to see if the tool is evaluating the same areas of interest they have themselves identified. This helps clinicians gauge plausibility and reliability. With novel and emerging diseases, like monkeypox as well as others we expect to increasingly see in the future, clinicians have little specific past clinical experience to rely on, but would identify areas of interest on radiologic imaging or dermatologic exams and it will be reassuring that the tool is focusing on the same area. This will help clinicians understand what may lead to false negative or positive results, especially if they need to override the tool's determinations. This will allow tools to be deployed more effectively in early outbreaks with an extra layer of oversight, especially as outbreaks may result in different clinical findings, whether radiologic or dermatologic, from the training datasets, as the outbreak expands. We often see as outbreaks spread beyond initial groups (whether due to geographic spread or due to different risk factors, such as we have seen with monkeypox) that clinical findings may change. Clinicians may also need to differentiate a new pathogen from other diseases as an outbreak spreads, given diseases prevalence varies by geography and between populations, and so monitoring how the tool works with different diseases will help. A lightweight tool that could provide an additional layer of oversight for new tools would help harness pattern recognition more effectively in evolving outbreaks. Inspired by the work of \citep{collins2018deep}, we introduce NMFx, a lightweight and general framework for analyzing activations based on nonnegative matrix decomposition (NMF) using no concept supervision or using image labels. We show that jointly factoring images in the feature space of a classification neural network allows extracting information about localization properties of the network without any image-level category annotations.

\bibliography{newbib}
\clearpage
\section*{Appendix}

\subsection{Related Work}
\paragraph{Visual Explanation.}
There have been a number of recent methods~\citep{das2020opportunities} whose goal is to visualize the learned representations of convolutional neural networks (CNNs) and explain their properties. Early techniques such as Zeiler and Fergus~\citep{zeiler2014visualizing} proposed deconvolution networks to identify parts of the input image that activate each neuron unit. Guided backpropagation ~\citep{springenberg2014striving} evaluates the effect of each neuron with respect to the input. \citep{yosinski2015understanding} introduces an optimization technique to synthesize images that highly activate specific neurons. \citep{zhou2016learning} shows that CNN layers act as unsupervised object detectors, and introduce class activation maps (CAM), a technique that uses the global average pooling (GAP) layer to generate importance heat maps for each class. This technique is limited to CNNs with a GAP layer and requires training of linear classifiers on top of the original networks. To address these limitations, Grad-CAM~\citep{selvaraju2017grad} and a number of other approaches~\citep{selvaraju2017grad,chattopadhay2018grad} have been developed, in particular, to extend CAM to other architectures. 

\paragraph{XAI Classification and Evaluation.}
As discussed in \citep{posadamoreno2022eclad}, XAI techniques can be grouped into two classes: local and global explanation techniques. \textit{Local methods}, also known as feature attribution techniques, seek to analyze and explain model behavior on a single data point (e.g., image). On the other hand, \textit{global methods} seek to analyze model behavior on a group of images and extract a set of representative concepts that would explain information learned by the models. A \textit{concept} can be any high level explanation, e.g., an object part or a super-pixel. Concepts can be provided by the user~\citep{kim2018interpretability} or extracted automatically~\citep{ghorbani2019towards}. Automatic methods first identify a set of concepts and then use optimization to determine their importance. For example, \citep{oramas2019visual} solves a Lasso problem to identify which which CNN activations are important for specific classes, later relying on guided back propagation~\citep{springenberg2014striving} to propagate activations and generate visual explanations. \citep{ghorbani2019towards,posadamoreno2022eclad} use TCAV~\citep{kim2018interpretability} to identify top concepts. 

Finally, there exist a number of techniques for evaluating visual explanation and neural interpretations methods~\citep{samek2016evaluating,yang2019bim,hooker2019benchmark,oramas2019visual}, in part due to the ambiguity of discriminative features between the classes. Please see \citep{bodria2021benchmarking} for a survey. Since the goal of our work is to visualize concepts learned by disease classification networks, we ask a board-certified doctor to analyze and comment on the information encoded in the concepts. 

\paragraph{XAI for Medical Image Analysis.}
Nowadays, researchers in medical imaging are increasingly using XAI to explain the results of their AI algorithms. Van der Velden et al. summarize related XAI research in their survey~\citep{van2022explainable}, finding that most of the XAI papers in the medical analysis used post-hoc explanations as contrasted with model-based explanations, which means the explanation was provided on a deep neural network (DNN) that had already been trained instead of being incorporated in the training process~\citep{van2022explainable}. Mainstream XAI methods for medical data include CAM~\citep{khakzar2019learning}, Grad-CAM~\citep{chen2019lesion}, local interpretable model-agnostic explanations (LIME)~\citep{rajaraman2019visualizing}, layer-wise relevance propagation (LRP)~\citep{bohle2019layer,hagele2020resolving}, and Shapley additive explanations (SHAP)~\citep{van2020volumetric}. The newest attribution-based methods, such as GSInquire and concept-based methods, such as TCAV, also show high performance in many new tasks~\citep{singh2020explainable,wang2020covid,posadamoreno2022eclad}. Additionally, \citep{fan2021interpretability} reviews applications of interpretability in medicine in a comprehensive taxonomy. Compared with the rapid development at the methodological level of single modality, evaluations on XAI for multi-modal medical imaging task is challenging and still immature~\citep{jin2022evaluating}. XAI in medical imaging analysis gives insight into how machine learning and neural networks can make AI-based clinical decisions more understandable and help medical doctors diagnose patients accurately.

\subsection{Additional Experimental Results}
\paragraph{Feature, Optimization, and Post-Processing Details} We evaluate the VGG-16~\citep{simonyan2014very} and the EfficientNet-B3~\cite{tan2019efficientnet} networks as feature extractors. In VGG-16, features are extracted after the ReLU following the last convolutional layer (layer 29). In EfficientNet-B3, features are extracted after the \verb|top_activation| layer. In each case, each objective is optimized until convergence. For all experiments, the default number of topics is the total number of classes, unless otherwise specified. All experiments were performed in Python, on a single node in a cluster with 200 GB memory and a 32GB GPU. Implementation details of each method are described below.

\paragraph{Classification Accuracy}
For reference, in Table~\ref{tab:accuracy}, we list the test accuracies of the neural networks trained for both classification tasks.
\begin{table}[]
\center
\begin{tabular}{|l|l|l|}
\hline
         & Monkeypox                   & Tuberculosis                      \\ \hline
Accuracy & \multicolumn{1}{c|}{0.8864} & \multicolumn{1}{c|}{0.9837} \\ \hline
\end{tabular}
\caption{Classification accuracy of neural networks on two considered medical classification tasks.}
\label{tab:accuracy}
\end{table}

\begin{figure}[]
\floatconts
  {fig:vis_monkeypox_nmfx_analysis}
  {\caption{Examples of predictions in the monkeypox classification tasks with corresponding NMFx topics.}}
  {\setlength{\jmlrminsubcaptionwidth}{0.95\linewidth}%
    \subfigure[\small Correctly classified monkeypox (TP).][c]{\label{fig:image-a3}%
     \includegraphics[width=1.0\linewidth]{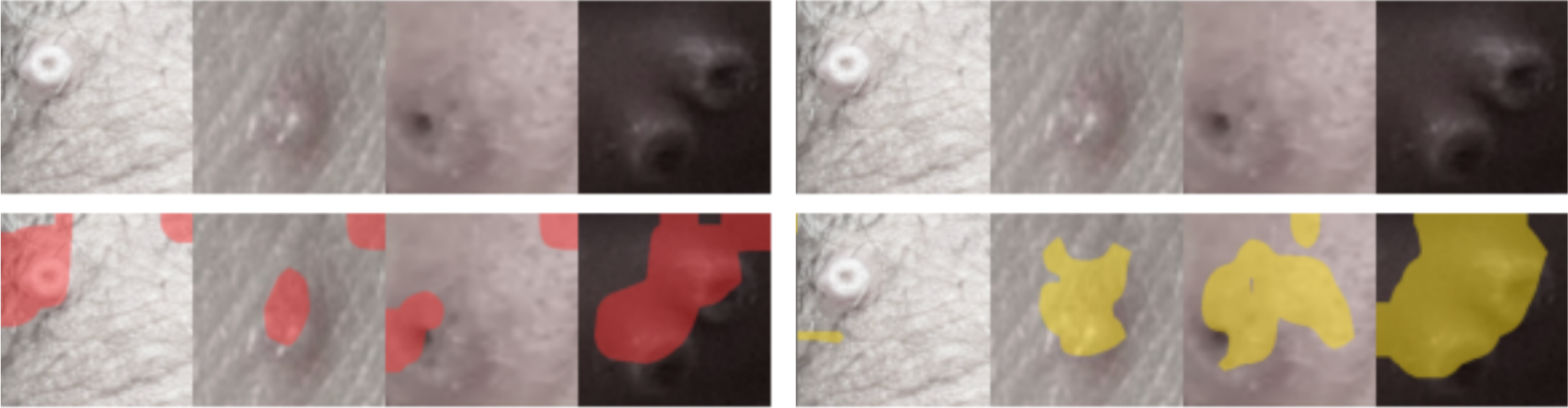}}%
    \qquad
    \subfigure[\small Correctly classified non-monkeypox (TN).][c]{\label{fig:image-b3}%
      \includegraphics[width=0.95\linewidth]{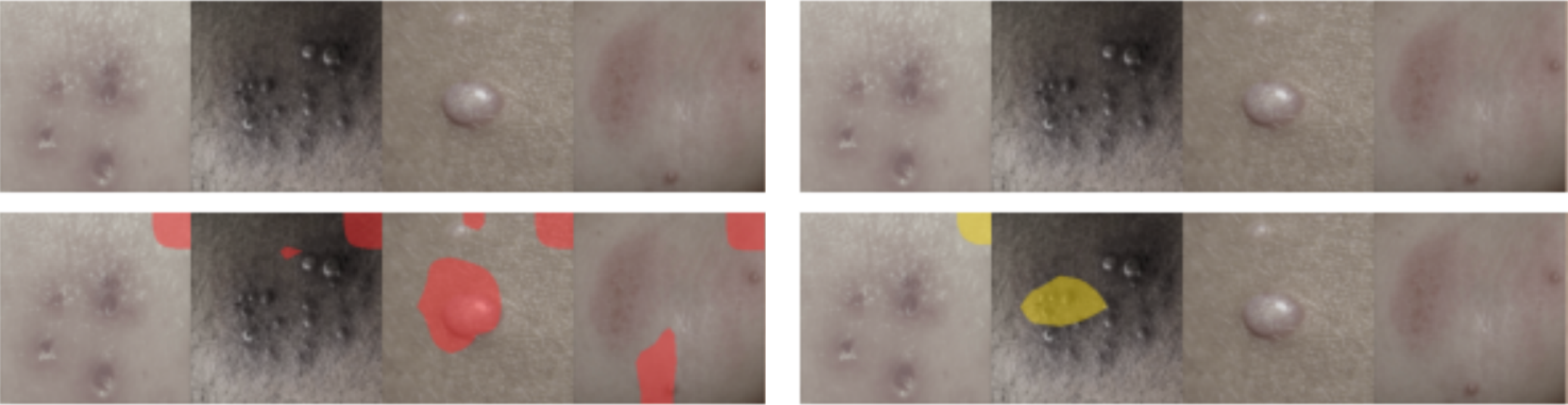}}
    \qquad
    \subfigure[\small False predictions of monkeypox (FP).][c]{\label{fig:image-b3}%
      \includegraphics[width=0.95\linewidth]{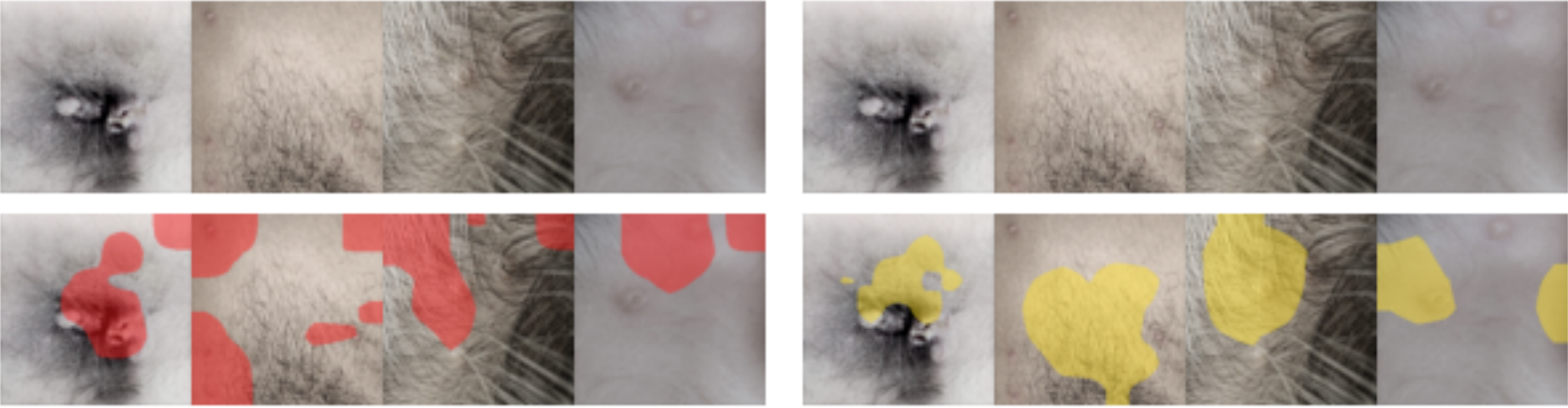}}
    \qquad
    \subfigure[\small False predictions of non-monkeypox (FN).][c]{\label{fig:image-b3}%
      \includegraphics[width=0.95\linewidth]{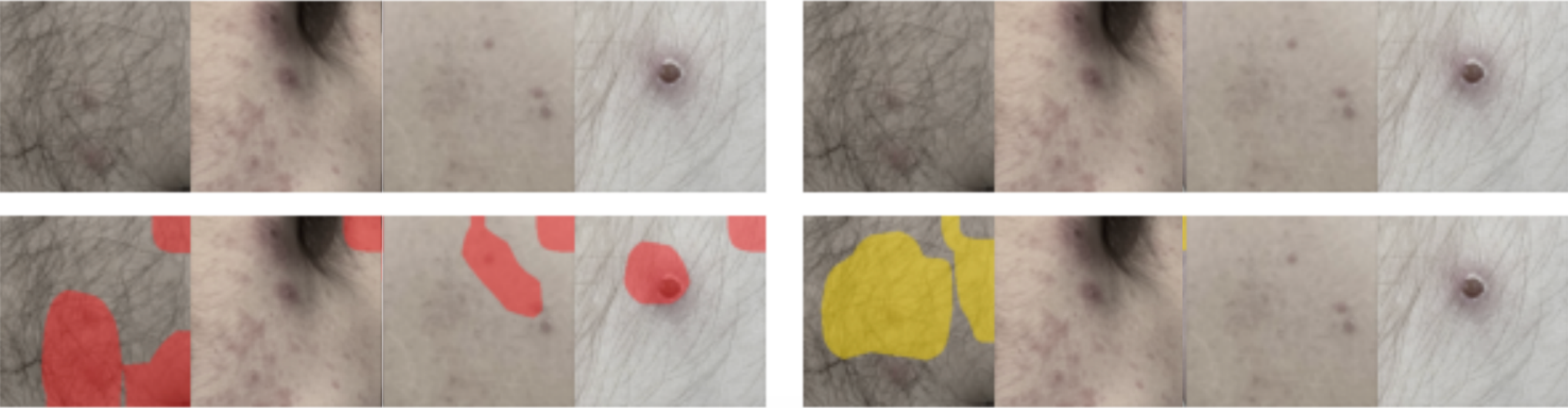}}
  }%
\end{figure}

\paragraph{Ablation Study: Prediction Analysis}

In Figure~\ref{fig:vis_monkeypox_nmfx_analysis}, we analyze examples that correctly and incorrectly classified and study their corresponding topics. We find that the model correctly classifies examples (both monkeypox and non-monkeypox) which are prototypical examples of disease. In these cases, the NMFx topics also consistently identify areas corresponding to disease. On the other hand, incorrect predictions correspond to out of distribution images. In these cases, the visual topics are scattered and not consistently identifying a particular region.

\paragraph{Ablation Study: Effect of Label Supervision and Number of Topics}

\begin{figure}[]
    \centering
    \includegraphics[width=1.0\linewidth]{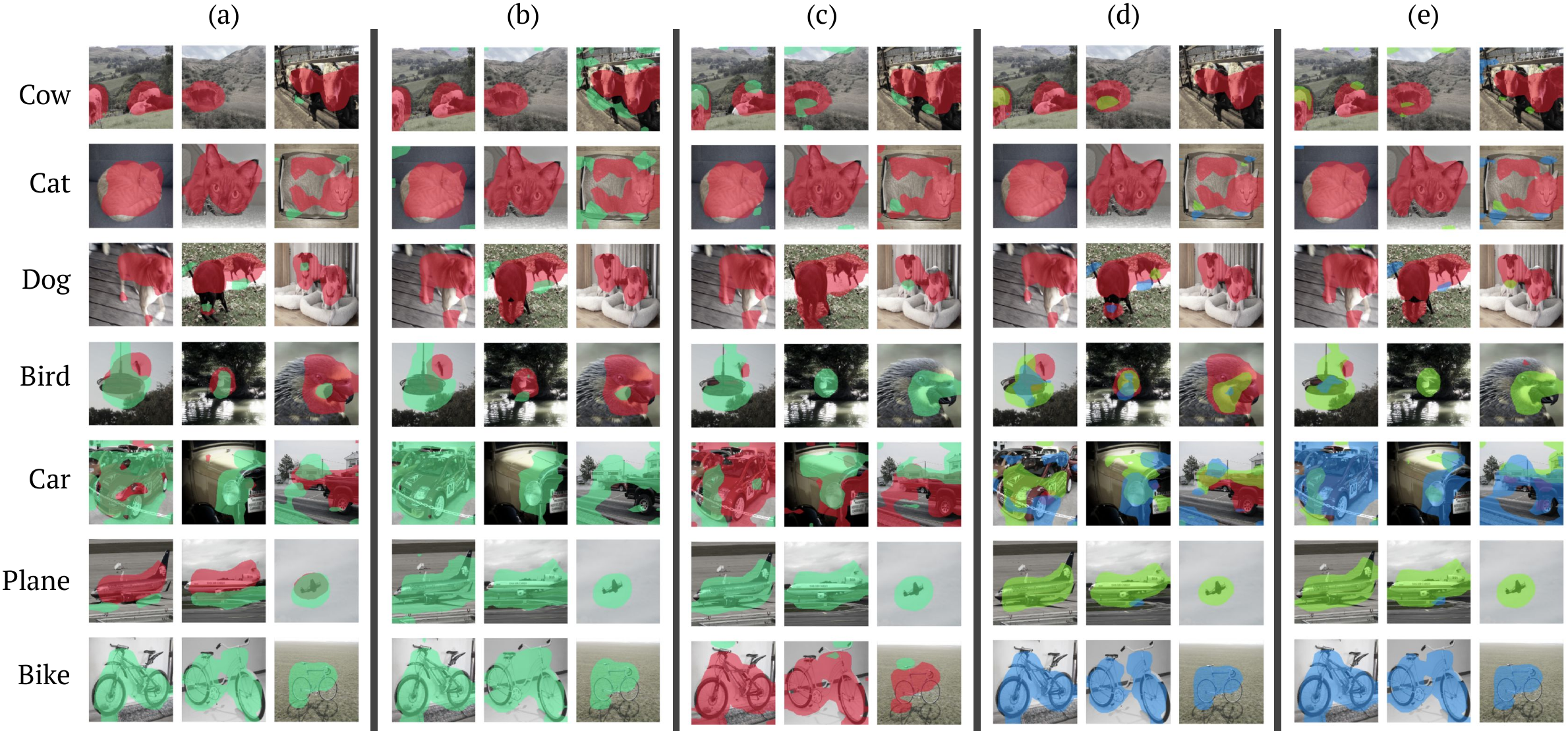}
    \caption{\emph{Visualization of heat maps produced by both unsupervised NMFx and supervised NMFx on a subset of images from the PASCAL-VOC 2010, as we vary the provided labels to demonstrate their impact on the heat maps. In (a) and (d), we run unsupervised NMFx with $K=2$ and $K=3$ topics, respectively. In (b) and (c), we report results of (SS)NMFx with $K=2$ topics and trained with class labels that separate (b) animals from vehicles and (c) flying objects from non-flying objects. In (e), we run (SS)NMFx with $K=3$ topics and provided class labels that separate land animals, land vehicles, and flying objects.}} 
    \label{fig:label_figure}
\end{figure}

We can use the supervised version of NMFx to specify labels that the input images belong to, and visually check whether the extracted topics can be grouped into labels. To demonstrate the ability of supervision to impact the heat maps produced by our method, we apply both unsupervised NMFx and supervised NMFx on a subset of images from seven classes (cow, cat, dog, bird, car, aeroplane, and bicycle) of the PASCAL-VOC 2010~\citep{pascal-voc-2010} data set. We find that our method works well for both black and white and color images, in particular, handling the larger variation in objects and their backgrounds found in PASCAL-VOC 2010. In Figure~\ref{fig:label_figure}, we display the heat maps produced for test images resulting from running unsupervised NMFx at (a) $K=2$ and (d) $K=3$, as well as supervised NMFx with three supervision labeling. In, (b) we use K=2 with two classes: animals (cow, cat, dog, bird) and vehicles (car, aeroplane, bicycle). In (c), we use $K=2$ with two classes: flying objects (bird, aeroplane) and non-flying objects (cow, cat, dog, car, bicycle). In (e), we use $K=3$ with three classes: land animals (cow, dog, cat), flying objects (bird, aeroplane), and land vehicles (car, bicycle). In each case, adding supervision causes the heat maps to better match the provided class labels. For $K=2$, we see that each of the two groupings of the classes in (b) and (c) shift the heat maps in the desired direction from the heat maps produced in (a). Here, (c) especially shifts the heat maps to group together objects that were distinct in (a), such as grouping cows with bikes (non-flying objects) and grouping birds with aeroplanes (flying objects), to respect the provided labels. Similarly, for $K=3$, the supervision leads to bikes and cars (land vehicles) being grouped together. This suggests that with supervision, our method is able to identify class-distinct features. We have also compared NMF and SSNMF results on the medical image applications discussed earlier, however, we did not observe consistently different performance.

\paragraph{Comparison to Other Techniques}
The first method we compare to is the method of \citep{oramas2019visual}, which follows a two step approach: first, a Lasso optimization problem is set up to identify the importance of features to the class of interest. Second, once filters that are important for prediction a certain class are identified, guided back-propagation~\citep{springenberg2014striving} is applied to generate a saliency heatmap. In comparison, our method generates heatmaps in a single step, and does not require backpropagation. For a fair comparison, we re-implement the method of \citep{oramas2019visual} and compare to our approach using the same network architecture and layers. The second method we compare to is ECLAD~\citep{posadamoreno2022eclad}, a recent method for automatic explainable concept extraction. Similar to \citep{ghorbani2019towards}, ECLAD automatically finds concepts, but additionally locates them within the image. The underlying algorithm rescales multiple levels of the activation map and analyzes each level at the pixel level. We use the default released implementation of ECLAD with the VGG-16 backbone.

%\bibliography{newbib}

\end{document}